\documentclass{article}

\usepackage{microtype}
\usepackage{graphicx}
\usepackage{booktabs} 

\usepackage{subcaption}
\usepackage{xcolor}
\usepackage{todonotes}

\usepackage{amsmath,amsfonts,bm}

\def\eqref#1{equation~\ref{#1}}

\def\1{\bm{1}}

\DeclareMathAlphabet{\mathsfit}{\encodingdefault}{\sfdefault}{m}{sl}
\SetMathAlphabet{\mathsfit}{bold}{\encodingdefault}{\sfdefault}{bx}{n}

\def\gD{{\mathcal{D}}}

\def\gL{{\mathcal{L}}}

\def\gO{{\mathcal{O}}}

\def\sR{{\mathbb{R}}}

\DeclareMathOperator*{\argmin}{arg\,min}

\definecolor{cadmiumgreen}{rgb}{0.0, 0.42, 0.24}

\newcommand{\best}[1]{\textbf{#1}}

\usepackage{hyperref}

\usepackage[accepted]{icml2021}

\icmltitlerunning{Multi-headed Neural Ensemble Search}

\begin{document}

\twocolumn[
\icmltitle{Multi-headed Neural Ensemble Search}



\icmlsetsymbol{equal}{*}

\begin{icmlauthorlist}
\icmlauthor{Ashwin Raaghav Narayanan}{fr}
\icmlauthor{Arbër Zela}{fr}
\icmlauthor{Tonmoy Saikia}{fr}
\icmlauthor{Thomas Brox}{fr}
\icmlauthor{Frank Hutter}{fr,bo}
\end{icmlauthorlist}

\icmlaffiliation{fr}{University of Freiburg}
\icmlaffiliation{bo}{Bosch Center for Artificial Intelligence}

\icmlcorrespondingauthor{Ashwin Raaghav Narayanan}{anarayan@cs.uni-freiburg.de}
\icmlcorrespondingauthor{Arbër Zela}{zelaa@cs.uni-freiburg.de}
\icmlcorrespondingauthor{Tonmoy Saikia}{saikiat@cs.uni-freiburg.de}

\icmlkeywords{Machine Learning, ICML}

\vskip 0.3in
]



\printAffiliationsAndNotice{}  

\begin{abstract}
    Ensembles of CNN models trained with different seeds (also known as Deep Ensembles) are known to achieve superior performance over a single copy of the CNN.
Neural Ensemble Search (NES) can further boost performance by adding architectural diversity.
However, 
the scope of NES remains prohibitive under limited computational resources.
In this work, we extend NES to multi-headed ensembles, which consist of a shared backbone attached to multiple prediction heads. 
Unlike Deep Ensembles, these multi-headed ensembles can be trained end to end, which enables us to leverage one-shot NAS methods to optimize an ensemble objective. With extensive empirical evaluations, we demonstrate that multi-headed ensemble search finds robust ensembles $3\times$ faster, while having comparable performance to other ensemble search methods, in both predictive performance and uncertainty calibration.


\end{abstract}


\section{Introduction}

Ensembles of neural networks are a common solution to improve predictive performance and uncertainty calibration \citep{hansen1990neural}. Ensembles of networks trained with different random initializations (known as \emph{Deep Ensembles}) can lead to well-performing and diverse models as they include different local optima \citep{lakshminarayanan2016simple, fort2019deep}.
They have also been shown to outperform approximate Bayesian methods \citep{lakshminarayanan2016simple, ovadia2019trust, gustafsson2020evaluating}.
On top of that, Neural Ensemble Search (NES)~\citep{zaidi2020neural} and HyperDeepEns~\citep{wenzel2021hyperparameter} can further improve performance by searching for a complementary set of ensemble members.
However, building and maintaining multiple models can be expensive.

While more efficient alternatives have been proposed for constructing ensembles, such as BatchEnsembles \citep{wen2020batchensemble} and Snapshot Ensembles \citep{huang2017snapshot,loshchilov2017sgdr}, they do not typically outperform deep ensembles. Multi-headed networks are another means to build efficient ensembles which uses a single shared backbone with multiple prediction heads \citep{lee2015m, lan2018knowledge}. 
While multi-headed models have been studied before, to the best of our knowledge, there is no prior work that studies the performance impact of searching head architectures.
In this work, we employ neural architecture search (NAS)~\citep{elsken2019neural} to search for the heads' architecture in the multi-headed models.
Unlike Deep Ensembles, multi-headed models are trained end-to-end, which allows one-shot NAS methods \citep{bender2018understanding, liu2019darts} to optimize an ensemble objective.

\begin{table}[t]
    \centering
    \caption{ Overview: Multi-headed NES (MH-NES) leads to performant models on CIFAR-100 with significantly less search costs. }
    \vspace{2mm}
	{
	\resizebox{1.0\linewidth}{!}{
	\begin{tabular}{lcccc}
	\toprule
	\textbf{Method} & \textbf{\# Params} & \textbf{Search Cost} & \textbf{Error} & \textbf{ECE} \\
					& \textbf{(M)} 			& \textbf{(GPU hours)} & & \\
	\midrule
	NES-RS & 4.0 & 15.2 & 19.75{\tiny$\pm$0.27} & 0.021{\tiny$\pm$0.001} \\
	HyperDeepEns (RS) & 4.2 & 28.7 & 20.11{\tiny$\pm$0.18} & 0.019{\tiny$\pm$0.003} \\
	MH-NES & \textbf{2.7} & \textbf{5.0} & \textbf{19.65{\tiny$\pm$0.09}} & \textbf{0.017{\tiny$\pm$0.001}} \\
	\bottomrule
	\end{tabular}
	}
	}
    \label{tab:EX:ens_cost_teaser}
\end{table}


The key contributions of this paper are as follows: 
\begin{itemize}
\itemsep 0pt\topsep-6pt\partopsep-6pt
    \item We extend the standard NAS search space to sample prediction heads in multi-headed ensembles. 
    \item We evaluate three one-shot NAS methods and demonstrate comparable or better performance to other ensemble search methods such as NES and HyperDeepEns, with significantly less search costs (see \autoref{tab:EX:ens_cost_teaser}).
    \item We provide an analysis and shed some light on potential factors affecting the performance of one-shot NAS methods in the multi-headed setting, particularly for larger ensemble sizes.
\end{itemize}

\subsection{Related Work}

\textbf{Ensemble} methods~\citep{hansen1990neural, dietterich2000ensemble} are extensively used in machine learning research, as both a means to boost performance, but also for better calibration and uncertainty estimates of deep neural networks~\citep{lakshminarayanan2016simple}.
Diversity in the predictions made by the ensemble base learners is believed to be crucial in order to obtain good ensembles. Some methods induce diversity by training with specialized losses \citep{lee2015m, zhou2018diverse}, building ensemble members with different topologies \citep{zaidi2020neural} or different training hyperparameters \citep{wenzel2021hyperparameter} to improve ensemble performance.
Multi-head networks trained under a unified objective can produce robust ensembles too, by utilizing diversity encouraging specialized losses \citep{lee2015m}, co-distillation \citep{lan2018knowledge} or both \citep{dvornik2019diversity}. 

\textbf{Neural Architecture Search} (NAS) aims to find an optimal neural network architecture that optimizes some objective (e.g. validation loss)~\citep{elsken2019neural}.
Early methods, such as reinforcement learning \citep{zoph2016neural} and evolutionary algorithms \citep{real2019regularized}, but also more recent efficient methods~\citep{bender2018understanding, elsken2017simple, cai2019proxylessnas, liu2019darts, xu2019pc, chen2021drnas} have shown that NAS can find architectures that can surpass hand-crafted ones.
Recently, NAS has also been applied to ensemble learning. Neural Ensemble Search (NES) \citep{zaidi2020neural} builds a pool of strong independent models with different architectures and applies ensemble selection to create the ensembles. On the other hand, AdaNAS~\citep{macko2019improving} iteratively adds base learners to an ensemble to improve the ensemble performance.
%
While NES and AdaNAS construct ensembles by independently training the base learners, which is computationally expensive, we focus on multi-headed ensembles that are cheaper to construct and allow us to leverage efficient NAS methods to design their topology.


\section{Background}

\subsection{Multi-headed Ensembles}

\begin{figure}[ht]
    \centering
    \includegraphics[width=0.8\linewidth]{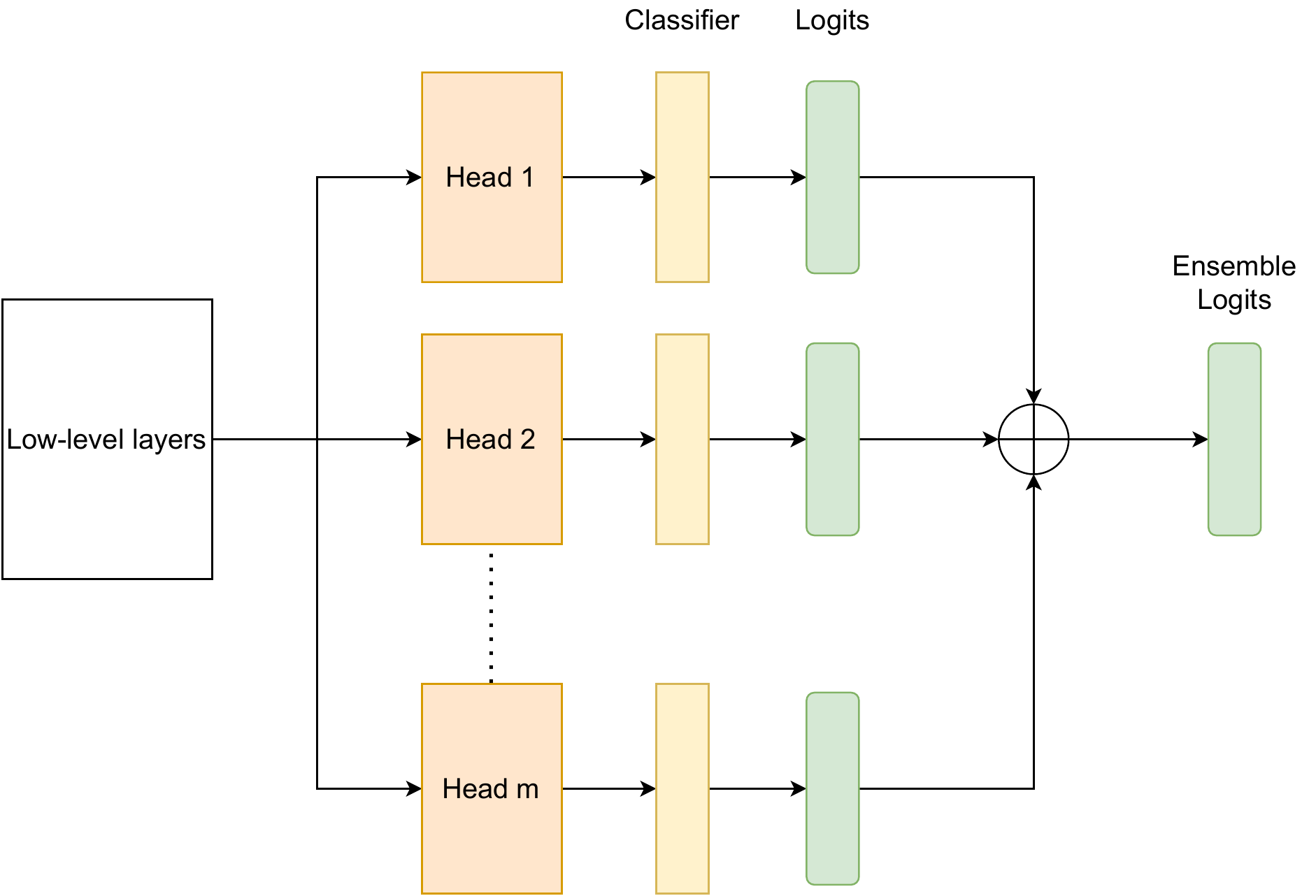}
    \caption{A multi-head ensemble for a CNN classifier}
    \label{fig:BG:mh_arch}
\end{figure}

Figure \ref{fig:BG:mh_arch} depicts an overview of a multi-headed ensemble for a classification task. The network can be divided into two sections: a \textit{lower block} shared by all the heads and \textit{multi-head block} consisting of $M$ heads. Each head may have similar or different architectures.

Given a classification task, let $\gD_{train} = \{(x_n, y_n) : n = 1,...,N \} $ be the training set, where $x_n \in \sR^D$ is the $D$-dimensional input and $y_n \in \{1,..., C\}$ is the label, assumed to be one of the $C$ classes.
We denote a base learner, in our case neural networks, by $f_\theta$, where $\theta$ represents the network parameters. A network takes the input $x_i$ and outputs a vector of probabilistic class posteriors over the classes as $f_\theta(x) \in \sR^C$.
We construct an ensemble $F$ of $M$ members $f_{\theta_1}, ..., f_{\theta_M}$ by averaging the output of the networks. 

Let $\ell$ be the cost function. 
We use an \emph{ensemble-aware loss function}, which includes the ensemble loss term, in addition to the individual head losses to improve the model performance:

\begin{equation}
    \gL_{train} = \sum_{i=1}^{M}{\ell(f_i(x), y)} + \ell(F(x), y).
    \label{eq:BG:mh_loss}
\end{equation}

\subsection{Differentiable Architecture Search}

Differentiable architecture search (DARTS)~\citep{liu2019darts} relaxes the discrete architecture space by assigning continuous architectural parameters to every operation choice in that space. Typically, the search is done for a cell (e.g., convolutional or recurrent) which is stacked in a repetitive manner to form the full network. The cell itself is represented as a directed acyclic graph (DAG) with an ordered sequence of $N$ nodes. Every node $x^{(i)}$ denotes a latent representation and each edge $(i, j)$ is associated with an operation $o^{(i,j)} \in \gO$ that transforms $x^{(i)}$, where $\gO$ is a pre-defined space of operations. 

The goal is to choose one operation from $\gO$ to connect each pair of nodes. 
DARTS continuously relaxes these discrete choices by creating a convex combination of the operations in  $\gO$ using the mixed operation $\bar{o}$: 
\begin{equation}
    \bar{o}^{(i,j)}(x) = \sum_{o \in \gO}{
        \frac{\exp(\alpha_o^{(i,j)})}{\sum_{o' \in \gO}{\exp(\alpha_{o'}^{(i,j)})}} o(x),
    }
    \label{eq:BG:mixed_ops}
\end{equation}
where $\alpha_o^{(i,j)}$ is the operation mixing weight. 
Thus, the problem of searching for the cell architecture boils down to learning a set of continuous variables $\alpha = \{\alpha^{(i,j)}\}$. This allows to formulate the NAS problem as a bi-level optimization problem:
\begin{equation*}
    \min_{\alpha}{\gL_{val}(w^*(\alpha), \alpha)}
    \quad
    \text{s.t.} \quad 
    w^*(\alpha) = \argmin_{w}{\gL_{train}(w, \alpha)}.
    \label{eq:BG:darts_bi}
\end{equation*}
After the search, 
the final architecture is derived by retaining the top-$k$ strongest operations from all the previous nodes.

In our work, we use three efficient variations of DARTS, namely:

\textbf{PC-DARTS} \citep{xu2019pc} introduces partial channel connections, where only a portion of channels are sent to the mixed operation.
To overcome the instability induced by sampling, it introduces edge weights, $\beta$, explicitly for every edge. The connectivity of an edge is then determined by combining both $\alpha$ and $\beta$.

\textbf{DrNAS} \citep{chen2021drnas} 
samples the $\alpha$ parameters from a parameterized Dirichlet distribution. 
Since it uses partial channel connections, it introduces a multi-stage scheme that increases the channel fraction while pruning the number of operators at every stage.

\textbf{RandomNAS} \citep{li2020random} builds a supernetwork similar to DARTS and simply uses randomly sampled architectures to train the shared weights during every mini-batch iteration. 
The trained shared weights are then used to evaluate the performance of multiple randomly sampled configurations and select the best performing one as the final architecture.


\section{NAS for multi-headed models}
\subsection{Search Space}

\begin{figure}
    \centering
    \includegraphics[width=0.7\linewidth]{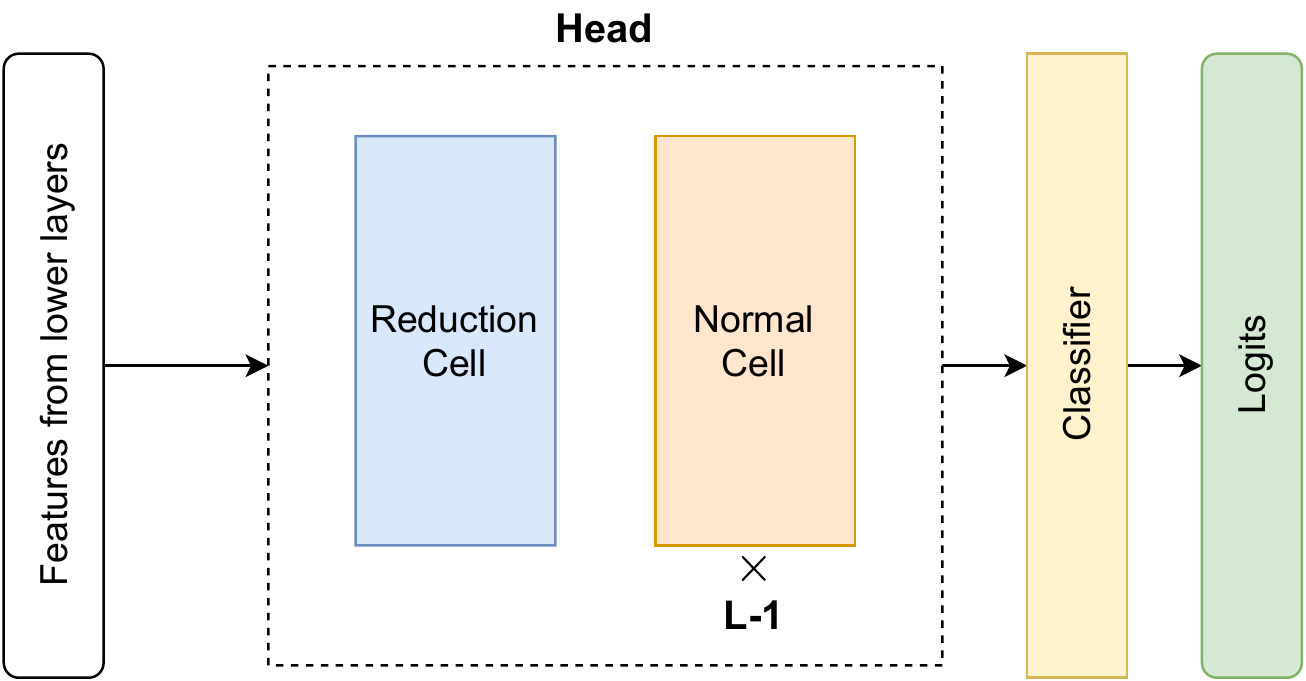}
    \caption{Single-headed network from the search space}
    \label{fig:AP:head_arch}
\end{figure}

Similar to the DARTS search space \citep{liu2019darts}, we use a cell-based architecture for every head, as shown in Figure \ref{fig:AP:head_arch}. A head is made of $L$ cells, with the first cell as a reduction cell (reducing spatial dimensions) and the remaining $L-1$ as normal cells (keeping spatial dimensions).
Unlike DARTS, we do not distinguish the configurations between reduction and normal cells. We found that having different configurations for reduction and normal cells had little to no impact in the performance of a multi-headed ensemble and only expanded the architecture space. 
Hence, in a multi-head ensemble with $M$ heads, there are $M \times L$ cells but only $M$ cell configurations or $M \times \{\alpha\}$ parameters to optimize.

\subsection{Diversity Encouraging Loss for Differentiable Search}

During the search phase of DARTS, there is no guarantee that the head architectures learnt via gradient descent would be diverse.
To this end, we introduce an additional diversity term only in the loss function for the architecture weights, $\gL_{val}$.
This ensures that the diversity in the one-shot model predictions originates from the architecture weights and not from the network weights.

We use the Jensen-Shannon Divergence (JSD) to measure the diversity between the individual head predictions and maximize it in the validation objective.
Unlike KL divergence, JSD is symmetric and bounded \citep{lin1991divergence}, which allows for direct maximization without the loss exploding.

Given $M$ heads, the diversity-encouraging loss term is:
\begin{align*}
    \gL_{jsd} &= JSD[f_{\theta_1}(x),...,f_{\theta_M}(x)] \\
            &= \frac{1}{M} \sum_{i=0}^{M}{
        KL[F(x) \parallel f_{\theta_i}(x)]
    } \\
    \gL_{val} &= \gL_{train} - \lambda_{jsd} \gL_{jsd}
    \label{eq:AP:archjsd}
\end{align*}

where $\lambda_{jsd}$ is the weight of the JSD loss. Refer to Appendix \ref{appendix:ablation} for ablations.

\section{Experiments} 

\subsection{Setup}
\textbf{Backbone architecture.} We use a WideResNet-40-2 \citep{zagoruyko2017wide} backbone for all the methods. The final block is then replaced with 3 cells from our search space.

\textbf{Baselines.} We compare our multi-headed ensemble methods with deep ensembles \citep{lakshminarayanan2016simple} (DeepEns (Sample) and DeepEns (RS)), NES-RS \citep{zaidi2020neural} and hyper-deep ensembles \citep{wenzel2021hyperparameter} (HyperDeepEns (RS)) from our search space. \textit{Sample} and \textit{RS} refer to head configurations that are randomly sampled and found by random search, respectively. We ran RS for 25 iterations (function evaluations). 

\textbf{Evaluation.} We compare the methods on NLL, classification error and ECE \citep{guo2017calibration}. All results are aggregates of 3 runs. We provide more details on the experimental setup in Appendix \ref{appendix:exp_details}. 

\subsection{Results}

\begin{figure}[ht]
    \centering
    \includegraphics[width=\linewidth]{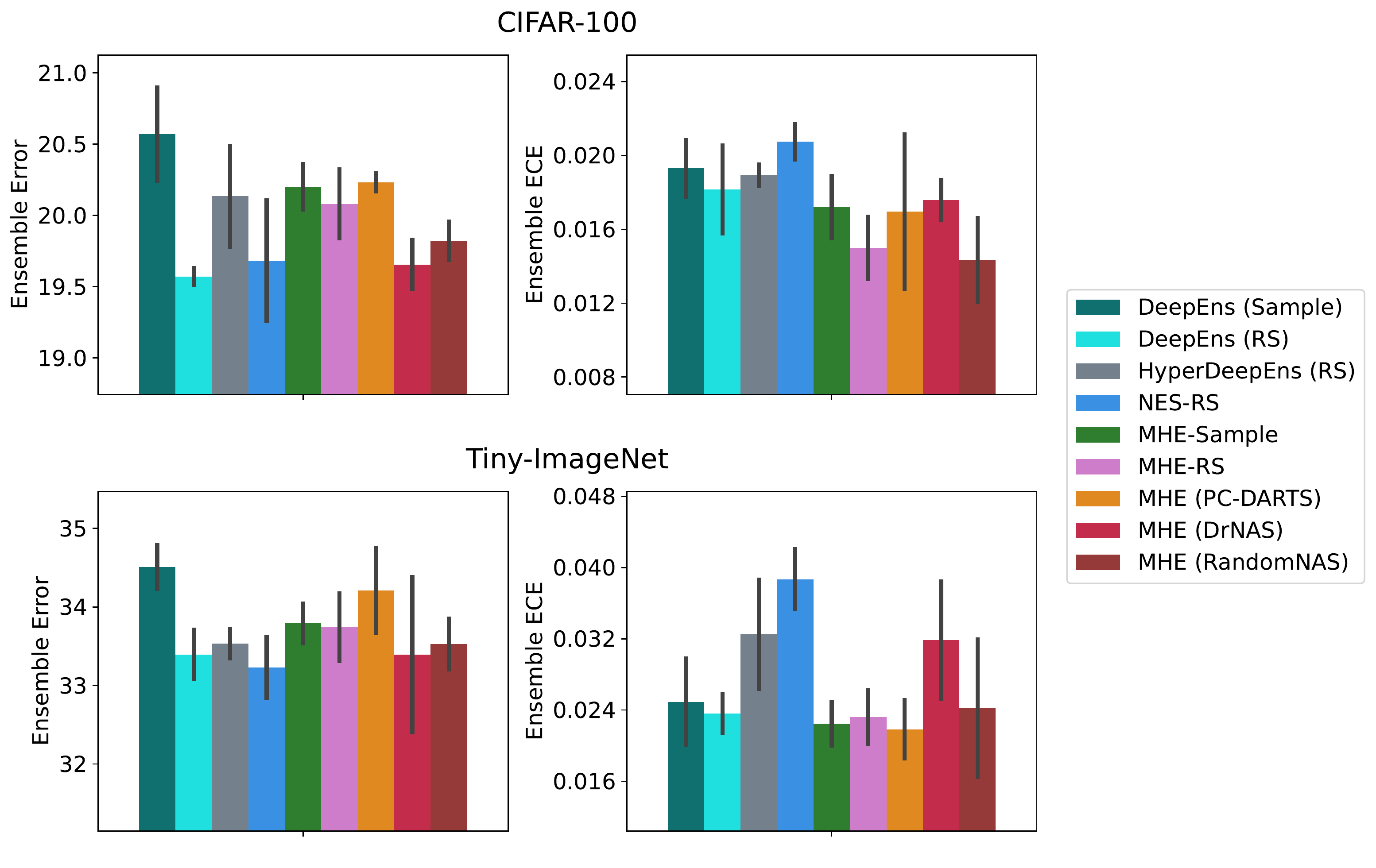}
    \caption{Ensemble performance comparison for $M=3$ (mean $\pm$ std.dev.). M is the number of ensemble members or prediction heads.
    }
    \label{fig:EX:ens_perf}
\end{figure}

\begin{figure}[ht]
    \centering
    \includegraphics[width=0.75\linewidth]{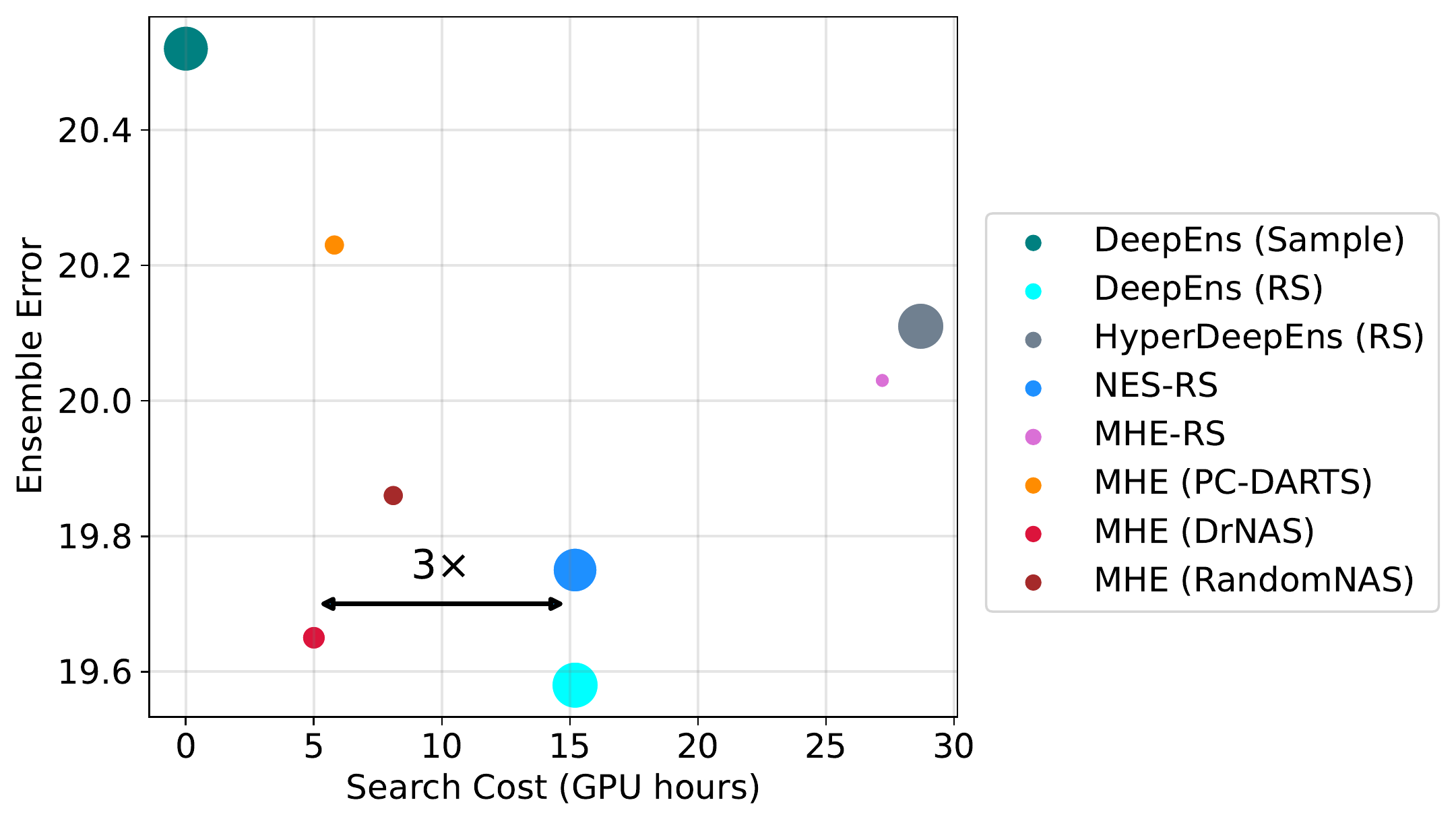}
    \caption{Ensemble error  vs search cost (for $M=3$) on CIFAR-100. Data points closer to the origin are better. Size of each point is proportional to the final ensemble's model size.}
    \label{fig:EX:ens_cost_c100}
\end{figure}

Figure \ref{fig:EX:ens_perf} shows the performance comparison of the ensemble methods on CIFAR-100 and Tiny-ImageNet for an ensemble size $M=3$. Figure \ref{fig:EX:ens_cost_c100} shows a comparison between the computational cost and ensemble performance.

As we can see, almost all multi-headed ensembles (even a randomly sampled configuration) show better uncertainty calibration than a deep ensemble.
Compared to NES and DeepEns (RS), the one-shot multi-headed search methods, especially DrNAS and RandomNAS, achieve comparable predictive performance, while being 3$\times$ more efficient during search.
Among the multi-headed methods, configurations identified by DrNAS perform the best in terms of prediction error. 
This can be attributed to the natural exploration-exploitation trade-offs in distribution learning. 
RandomNAS also performed almost as well for small ensembles and for CIFAR-100 identified better configurations than a naive and expensive random search.

Interestingly, for larger ensemble sizes the performance of differentiable methods deteriorates while methods based on random search achieve better performance (Appendix \ref{appendix:ens_perf}). 
We provide more insights into this behavior in the following section.

\subsection{Analysis of Differentiable Search Methods}
\label{EX:darts_analysis}

\citet{zela2020understanding} demonstrated that DARTS can often find degenerate architectures depending on the search space at hand.
Similarly, we found that PC-DARTS and DrNAS struggled to find architectures better than random search, or even random \emph{samples}, especially for larger ensemble sizes.

A Hessian norm analysis of $\nabla^2\gL_{valid}$ \citep{zela2020understanding} of these methods (Appendix \ref{appendix:darts_analysis}) indicated that this was not attributed to architecture over-fitting in the one-shot model: both PC-DARTS and DrNAS were able to find well performing configurations for a single network from the search space, though not for the multi-headed space.

\begin{figure}[ht]
    \centering
    \includegraphics[width=\linewidth]{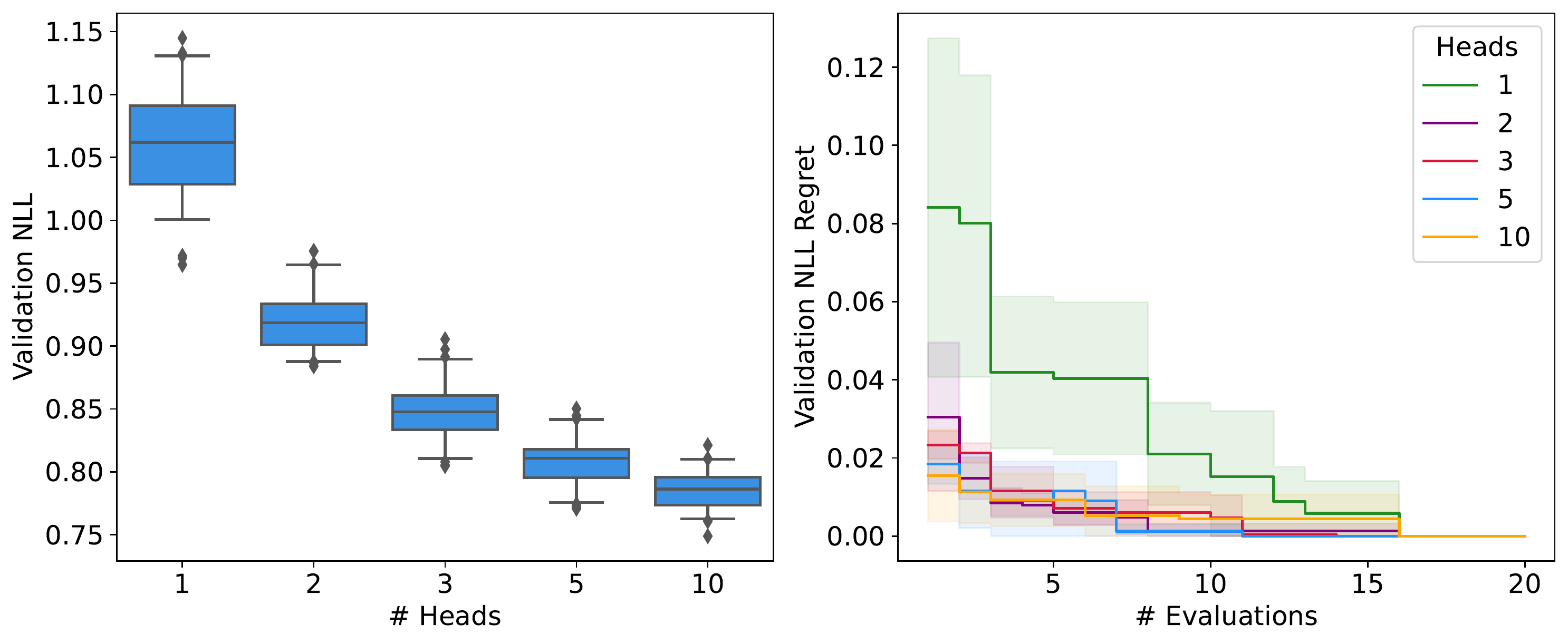}
    \caption{\textit{(left)} Distribution of validation NLL from 60 samples across 3 random search runs. \textit{(right)} Validation NLL Regret from optimal configuration of random search (mean $\pm$ std.dev.). All results on CIFAR-100.}
    \label{fig:EX:head_variance_c100}
\end{figure}

To this end, we analyze the search space by randomly sampling configurations for different ensemble sizes. Figure \ref{fig:EX:head_variance_c100} shows that while there is a clear improvement in the mean ensemble performance when scaling up the number of heads, the variance in the ensemble performance decreases significantly.
The validation regret curves flatten out with more heads, 
indicating that diminishing returns can be expected for running a more complex search method for $M > 5$ (while increasing computational demands).
A similar behaviour can be observed on Tiny-ImageNet in Appendix \ref{appendix:darts_analysis}.

Furthermore, 
while most NAS methods previously searched for only 2 cells~\citep{liu2019darts, zhou2019bayesnas, xu2019pc, chen2021drnas}, our search space size increases exponentially with the number of ensemble members/heads. We hypothesize that this, coupled with a low variance regime, leads the DARTS-based methods to get stuck in local optima and find suboptimal configurations.
This could also explain why random search traverses the operation search space better and identifies better performing configurations.


\section{Conclusion and Future Work} 

In this work, we introduced a method for efficiently searching for the architectures of the individual heads in a multi-headed ensemble.
The resulting models are accurate, well calibrated and also more efficient (computationally and memory-wise) compared to deep ensembles and other methods found using inefficient methods that require training every ensemble base learner individually. 
In the future, we would like to improve the robustness of differentiable NAS methods applied to ensemble learning.

{\small
\bibliographystyle{icml2021}
\bibliography{bib/ensemble,bib/nas,bib/misc}
}

\clearpage
\appendix

\section{Experiment Details}
\label{appendix:exp_details}

The models were implemented with a WideResNet-40-2 \citep{zagoruyko2017wide} backbone. The final block of the WideResNet is replaced with heads from the search space. 
All algorithms were implemented in PyTorch 1.7.1 and run on NVIDIA RTX 2080 GPUs. All the final evaluation networks were trained for 100 epochs with mini-batch size of 128 using SGD with initial learning rate $\eta=0.1$, momentum 0.9 and weight decay $3\times10^{-4}$. The learning rate was decayed by a cosine annealing schedule \citep{loshchilov2017sgdr} towards 0.

The one-shot model of the differentiable search methods is trained for 50 epochs with a batch size of 64 using SGD with the same optimizer settings as the final evaluation for the network parameters. The architecture parameters were initialized uniformly around zero. We used Adam optimizer \citep{kingma2017adam} with initial learning rate $\eta=3\times10^{-4}, \beta_1=0.5, \beta_2=0.999$ and weight decay $10^{-3}$. Similar to \cite{liu2019darts}, learnable affine parameters were disabled during search to avoid rescaling the outputs of the candidate operations. 
In PC-DARTS, the partial channel parameter $K$ was set to 4. For DrNAS, we use a two stage progressive learning approach, 25 iterations each. In in the first stage, the partial channel parameter $K=4$ and all 7 operations are used. For the second stage, half the candidate operations are pruned (i.e., 4 remain) and the network is widened by increasing $K$ to 2.
To avoid bias towards non-parametric operations, we warmstart the one-shot model by not updating the architecture parameters for the first 15 epochs for PC-DARTS and 10 epochs for each stage in DrNAS, following the original implementation.

All experiments are an aggregation of 3 independent runs and the results show the mean and standard deviation, unless specified otherwise. For search methods, an independent run includes one search and one evaluation phase.

\subsection{Baselines}

\begin{itemize}
    \item \textbf{DeepEns (Sample)} - Deep Ensembles \citep{lakshminarayanan2016simple} train $M$ different initializations of a randomly sampled head ($M=1$) from the search space.
    \item \textbf{DeepEns (RS)} is a deep ensemble built using a the optimal configuration from random search for $M=1$.
    \item \textbf{NES-RS} - The ensemble pool is constructed by sampling single-headed networks from our search space. Then, we run forward selection on the pool to select the ensemble members based on the validation set. 
    This approach can be seen as an adaptation of Neural Ensemble Search \citep{zaidi2020neural} to just the last block. 
    We sampled 25 single-headed configurations to build the ensemble pool.
    \item \textbf{HyperDeepEns (RS)} - Hyperparameter Ensembles \citep{wenzel2021hyperparameter} use the same networks as \textit{DeepEns (RS)} but train them with two varying hyperparameters: label smoothing and weight decay along with random initializations to build the ensemble pool. Forward selection is used to select the final ensemble members. We sampled 25 configurations to build the ensemble pool.
    \item \textbf{MHE-Sample} is a random sample from the multi-head search space.
    \item \textbf{MHE-RS} is the final configuration from random search. We randomly sample 25 multi-headed configurations from the search space and choose the best configuration based on the validation NLL.
\end{itemize}

\subsection{Hyperparameter optimization of search parameters}

Since this is a fairly new domain for differentiable search methods, we optimize the hyperparameters of the search phase using BOHB \citep{falkner2018bohb}, a multi-fidelity Bayesian optimization method that uses an efficient surrogate model to search for good hyperparameter configurations. 

The hyperparameters optimized were: network learning rate, network weight decay, architecture learning rate, architecture weight decay, backbone layers and width.
We included the number of backbone layers and widening factor to search for a proxy one-shot model that can search faster without compromising on the optimal configuration performance. 
The incumbent configuration from BOHB was later used to run PC-DARTS and DrNAS for all multi-headed settings.


\section{Additional Results}
\label{appendix:ens_perf}

\begin{table}[ht]
    \centering
    \caption{Performance and computation cost of different search methods for ensemble size $M=3$ on CIFAR-100.
    }
    \vspace{2mm}
	{
	\resizebox{1.0\linewidth}{!}{
	\begin{tabular}{lcccc}
	\toprule
	\textbf{Method} & \textbf{\# Params} & \textbf{Search} & \textbf{Error} & \textbf{ECE} \\
					& \textbf{(M)}       & \textbf{(GPU hours)} & & \\
	\midrule
	DeepEns (Sample) & 4.1 & & 20.52\tiny{$\pm$0.12} & 0.019\tiny{$\pm$0.001} \\
	DeepEns (RS) & 4.2 & 15.2 & 19.58\tiny{$\pm$0.03} & 0.018\tiny{$\pm$0.001} \\
	HyperDeepEns (RS) & 4.2 & 28.7 & 20.11\tiny{$\pm$0.18} & 0.019\tiny{$\pm$0.003} \\
	NES-RS & 4.0 & 15.2 & 19.75\tiny{$\pm$0.27} & 0.021\tiny{$\pm$0.001} \\
	MHE-RS & 2.4 & 27.2 & 20.03\tiny{$\pm$0.12} & 0.015\tiny{$\pm$0.001} \\
	MHE (PC-DARTS) & 2.6 & 5.8 & 20.23\tiny{$\pm$0.04} & 0.017\tiny{$\pm$0.002} \\
	MHE (DrNAS) & 2.7 & 5.0 & 19.65\tiny{$\pm$0.09} & 0.017\tiny{$\pm$0.001} \\
	MHE (RandomNAS) & 2.6 & 8.1 & 19.86\tiny{$\pm$0.08} & 0.015\tiny{$\pm$0.001} \\
	\bottomrule
	\end{tabular}
	}
	}
    \label{tab:EX:ens_cost}
\end{table}

\subsection{Ensemble Performance for larger ensembles}

\begin{figure}[ht]
    \centering
    \includegraphics[width=\linewidth]{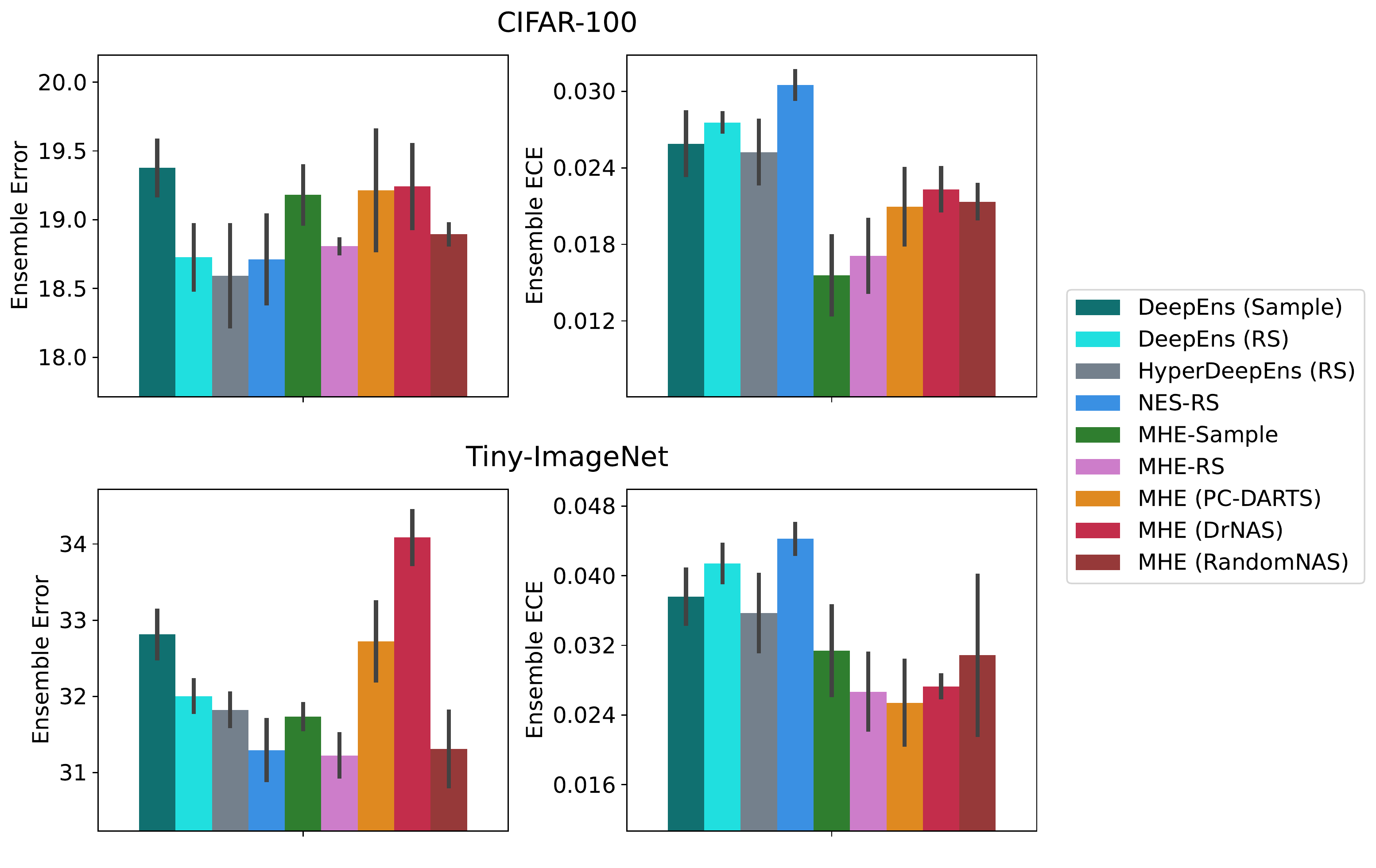}
    \caption{Ensemble performance comparison for $M=5$ (mean $\pm$ std.dev.)}
    \label{fig:appdx:ens_perf_m5}
\end{figure}

On larger ensembles, the multi-headed methods see significant improvements in ensemble performance and calibration, as shown in Figure \ref{fig:appdx:ens_perf_m5}. 
However, the configurations identified by DARTS-based methods, PC-DARTS and DrNAS, deteriorate and perform only as good as a random sample from the search space on CIFAR-100 and even worse on Tiny-ImageNet. 
The sub-optimal configurations can be attributed to the low variance regime and the increased search space size, which makes them get stuck in a local optima.
RandomNAS, the one-shot search method that is based on randomly sampled architectures, is more robust and achieves the best ensemble error for $M=5$.

\subsection{Ensemble Performance under Dataset Shift}

\begin{figure}[ht]
    \centering
    \includegraphics[width=\linewidth]{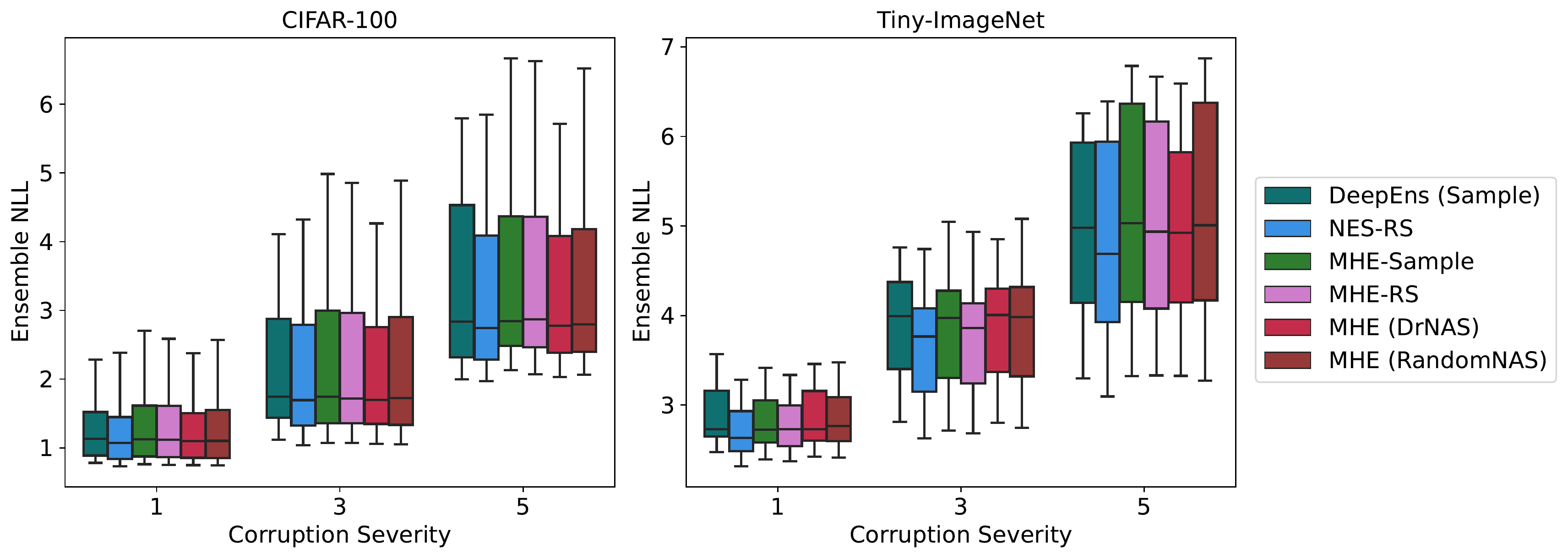}
    \caption{Ensemble performance comparison for $M=3$ across different shift severities}
    \label{fig:appdx:ens_shift}
\end{figure}

The results shown in Figure \ref{fig:appdx:ens_shift} shows how the ensembles perform on shifted data. The multi-headed search methods mostly outperform the deep ensemble baseline and are even on par with NES on CIFAR-100.


\section{Analysis of DARTS methods}
\label{appendix:darts_analysis}

\begin{table}[ht]
    \centering
    \caption{Performance of different search methods for one head on CIFAR-100 dataset. WideResNet block is the baseline block used in the original network.}
    
    {\scriptsize
    \begin{tabular}{ccc}
    \toprule
    \textbf{Method} & \textbf{NLL} & \textbf{Error} \\
    \midrule
    WideResNet block & 1.089\tiny{$\pm$0.006} & 24.41\tiny{$\pm$0.08} \\
    Random Sample & 0.983\tiny{$\pm$0.020} & 24.12\tiny{$\pm$0.21} \\
    Random Search & 0.927\tiny{$\pm$0.015} & \best{23.17}\tiny{$\pm$0.08} \\
    PC-DARTS & 0.923\tiny{$\pm$0.008} & \best{23.20}\tiny{$\pm$0.10} \\
    DrNAS & \best{0.908}\tiny{$\pm$0.007} & 23.61\tiny{$\pm$0.21} \\
    \bottomrule
    \end{tabular}
    }
    \label{tab:EX:perf_m1}
\end{table}

\begin{figure}[ht]
    \centering
    \begin{subfigure}[t]{\linewidth}
        \centering
        \includegraphics[width=0.8\linewidth]{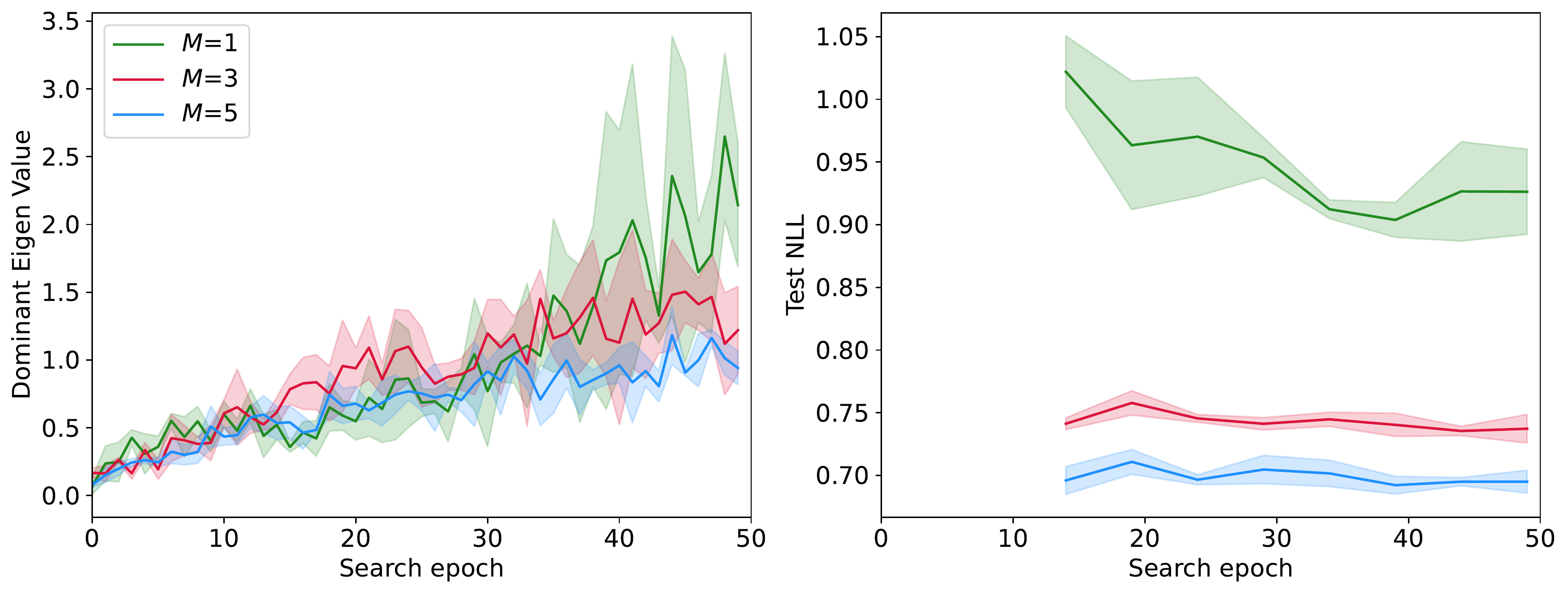}
        \caption{PC-DARTS}
    \end{subfigure}
    \begin{subfigure}[t]{\linewidth}
        \centering
        \includegraphics[width=0.8\linewidth]{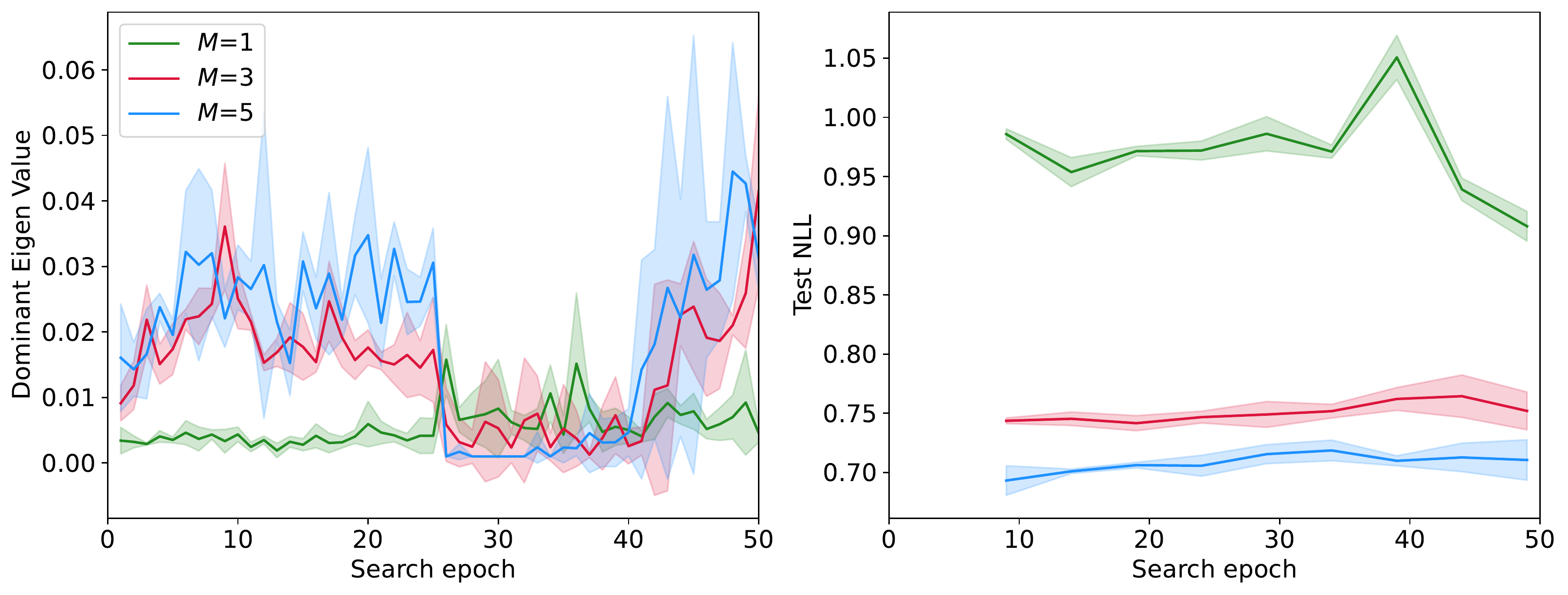}
        \caption{DrNAS}
    \end{subfigure}
    
    \caption{\textit{(left)} Trajectory of the Hessian norm i.e., dominant Eigenvalue of $\nabla^2\gL_{valid}$ for PC-DARTS and DrNAS. \textit{(right)} Test NLL of the optimal architectures by the search methods. All experiments conducted on CIFAR-100. }
    \label{fig:EX:eigval_analysis}
\end{figure}

Figure \ref{fig:EX:eigval_analysis} shows the performance of both gradient-based search methods PC-DARTS and DrNAS using the incumbent configuration. 
Following RobustDARTS \citep{zela2020understanding}, we plot the dominant eigenvalue of the Hessian of validation loss ($\nabla^2\gL_{valid}$), which serves as a proxy for sharpness, for each search epoch and the corresponding test NLL for 1, 3 and 5 heads.

Both PC-DARTS and DrNAS found good single-head configurations that performed at least as well as random search, as shown in Table \ref{tab:EX:perf_m1}. The DARTS search space is so tuned that random search can offer comparable performance to differentiable methods \citep{li2020random}. If we include early stopping based on the dominant eigenvalue like DARTS-ES \citep{zela2020understanding}, PC-DARTS produces an even better configuration.
However, for more heads, the differentiable methods fail to offer much more than a random sample on in-distribution performance. The Hessian norm indicates that this is not a result of architecture over-fitting in the one-shot model.

\begin{figure}[h]
    \centering
    \includegraphics[width=\linewidth]{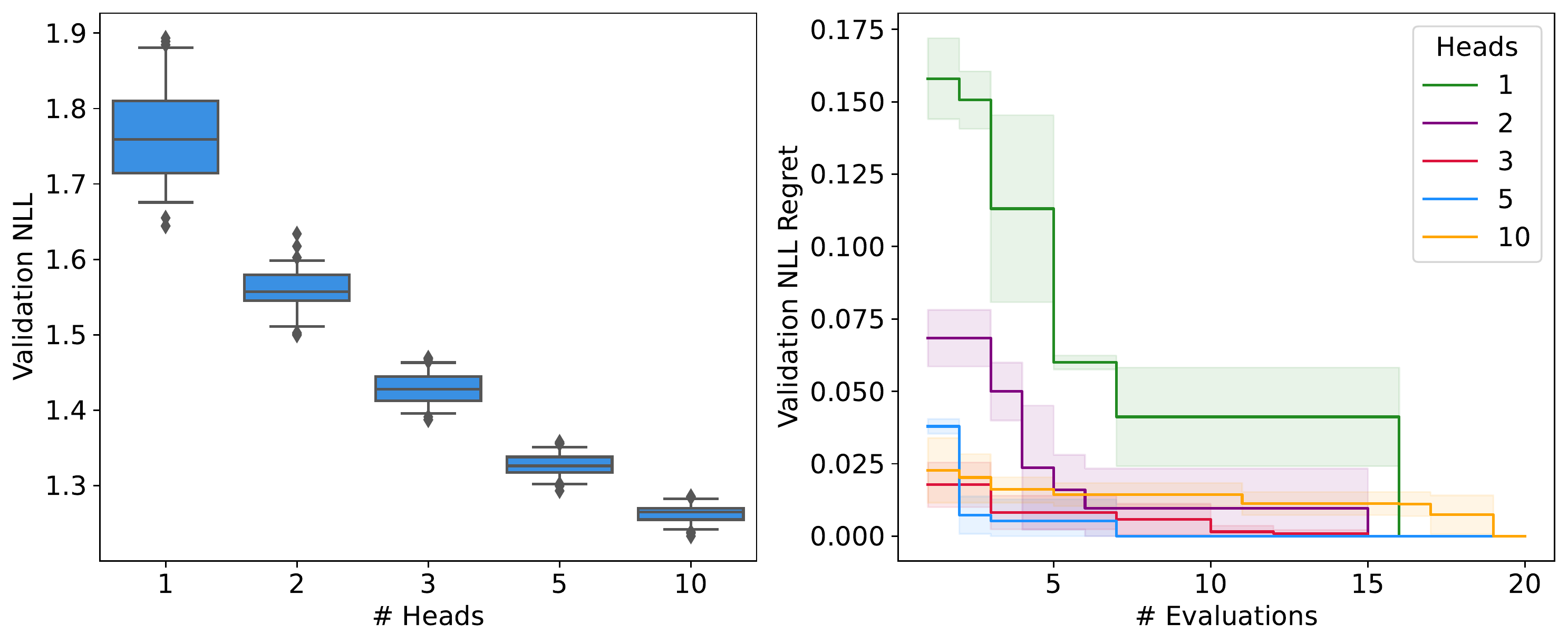}
    \caption{\textit{(left)} Distribution of validation NLL from 60 samples across 3 random search runs. \textit{(right)} Validation NLL Regret from optimal configuration of random search (mean $\pm$ std.dev.). All results on Tiny-ImageNet.}
    \label{fig:EX:head_variance_tim}
\end{figure}

\begin{figure}[ht]
    \centering
    \includegraphics[width=1\linewidth]{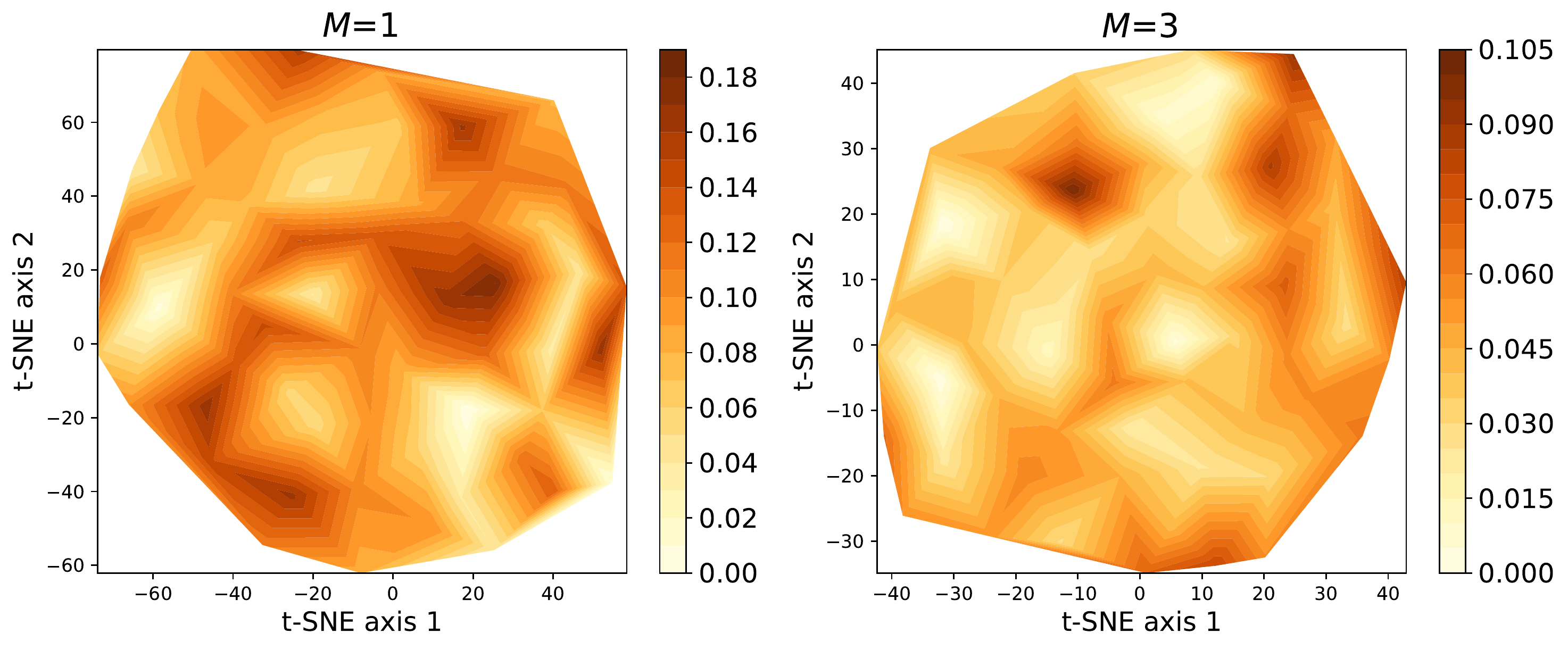}
    \caption{t-SNE visualization of 60 different configurations sampled randomly for ensemble sizes 1 and 3 on CIFAR-100. The regions are coloured based on the validation NLL Regret. Lighter colour indicates better performance. The pervasiveness of bright regions for ensembles implies more well-performing regions in the architecture space.}
    \label{fig:EX:arch_surface}
\end{figure}

We build a t-SNE \citep{van2008visualizing} visualization of the head architectures, as shown in Figure \ref{fig:EX:arch_surface} to visualize the validation NLL across the architecture search space.
We convert the architecture into a vector of edges (operators) for all heads and visualize them using t-SNE with hamming distance as the distance measure. The assumption is that cells that share similar edges would be closer to each other than cells with different operators and this should translate into their performance.

As we can see in Figure \ref{fig:EX:arch_surface}, the architecture space when $M=1$ is defined with bright and dark subspaces, indicating the clear distinction in the quality of architectures.
Moreover, the well-performing subspaces are fewer and far between, leading to more informative gradients.
Even for a small ensemble $M=3$, while there are still the occasional dark subspaces, the bright spaces become more pervasive, highlighting the similar performance levels across most configurations. The validation NLL regret is predominantly less than 0.04-0.06, which is 3$\times$ smaller than the regret in single heads.


\section{Ablation Studies}
\label{appendix:ablation}

\subsection{Diversity Term in Architecture Search}

\begin{figure}[ht]
    \centering
    \begin{subfigure}{\linewidth}
        \centering
        \includegraphics[width=1\linewidth]{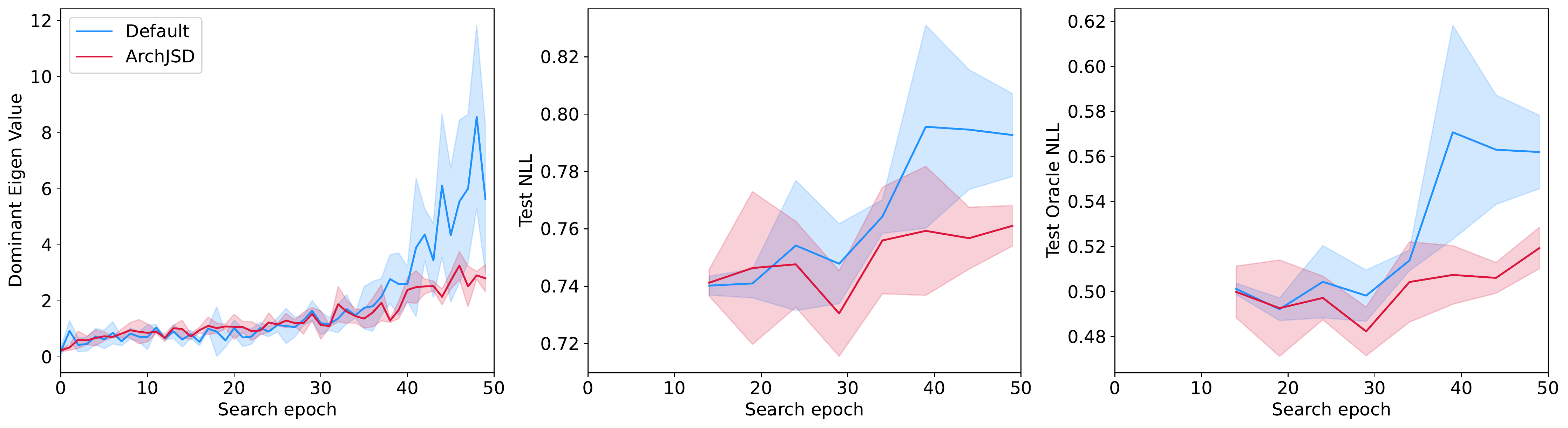}
        \caption{PC-DARTS}
    \end{subfigure}
    \begin{subfigure}{\linewidth}
        \centering
        \includegraphics[width=1\linewidth]{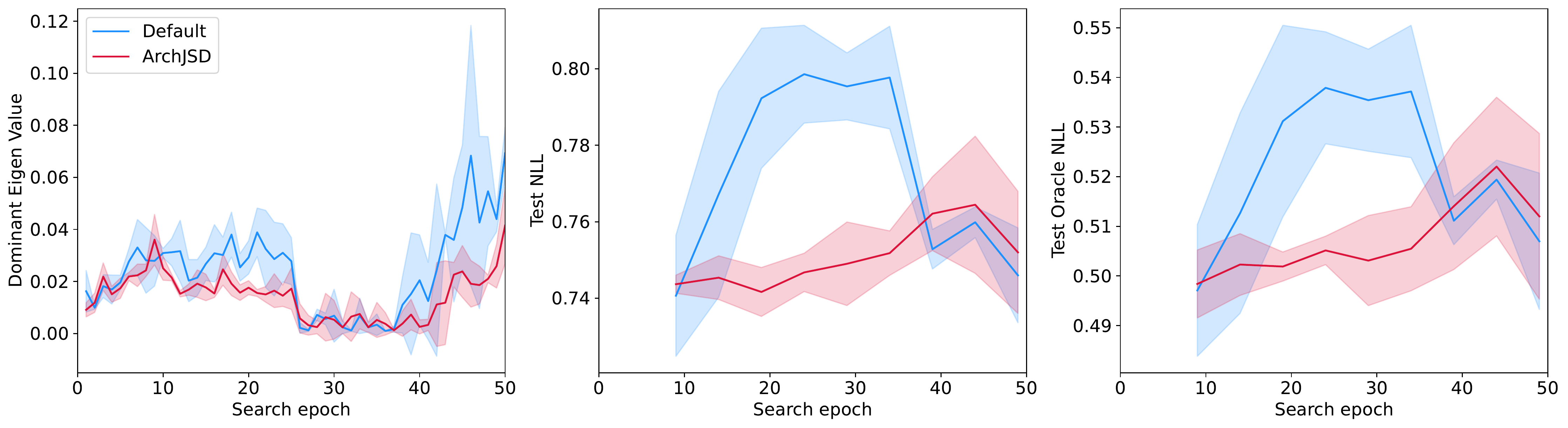}
        \caption{DrNAS}
    \end{subfigure}
    \caption{\textit{(left)} Trajectory of the Hessian norm (dominant eigenvalue of $\nabla^2\gL_{valid}$) during search. \textit{(middle)} Test NLL of optimal architectures from the search phase. \textit{(right)} Oracle Ensemble NLL \citep{zaidi2020neural} of the optimal architectures during search. All experiments run on CIFAR-100 with an ensemble size $M=3$. Default configurations were used for the search methods. (mean $\pm$ std.dev.)}
    \label{fig:appendix:archjsd_search}
\end{figure}

Figure \ref{fig:appendix:archjsd_search} compares the performance of PC-DARTS and DrNAS with and without the diversity constraint on the default search configurations, before hyperparameter optimization.
In PC-DARTS, it is clear that the additional constraint improves the performance of final configuration. The dominant eigenvalue of $\nabla^2\gL_{valid}$ is more regulated, indicating a more stable search.
The optimal configuration has a better test performance and diversity performance, demonstrating the benefits of using the diversity term during search.
There is no clear winner for DrNAS. While the diversity term regulated the dominant eigenvalues and led to better early configurations, the final configuration did not improve the diversity or accuracy in the predictions over the unconstrained objective.

\subsection{Deep Ensembles vs MHE for Similar Architectures}

\begin{figure}[ht]
    \centering
        \includegraphics[width=1\linewidth]{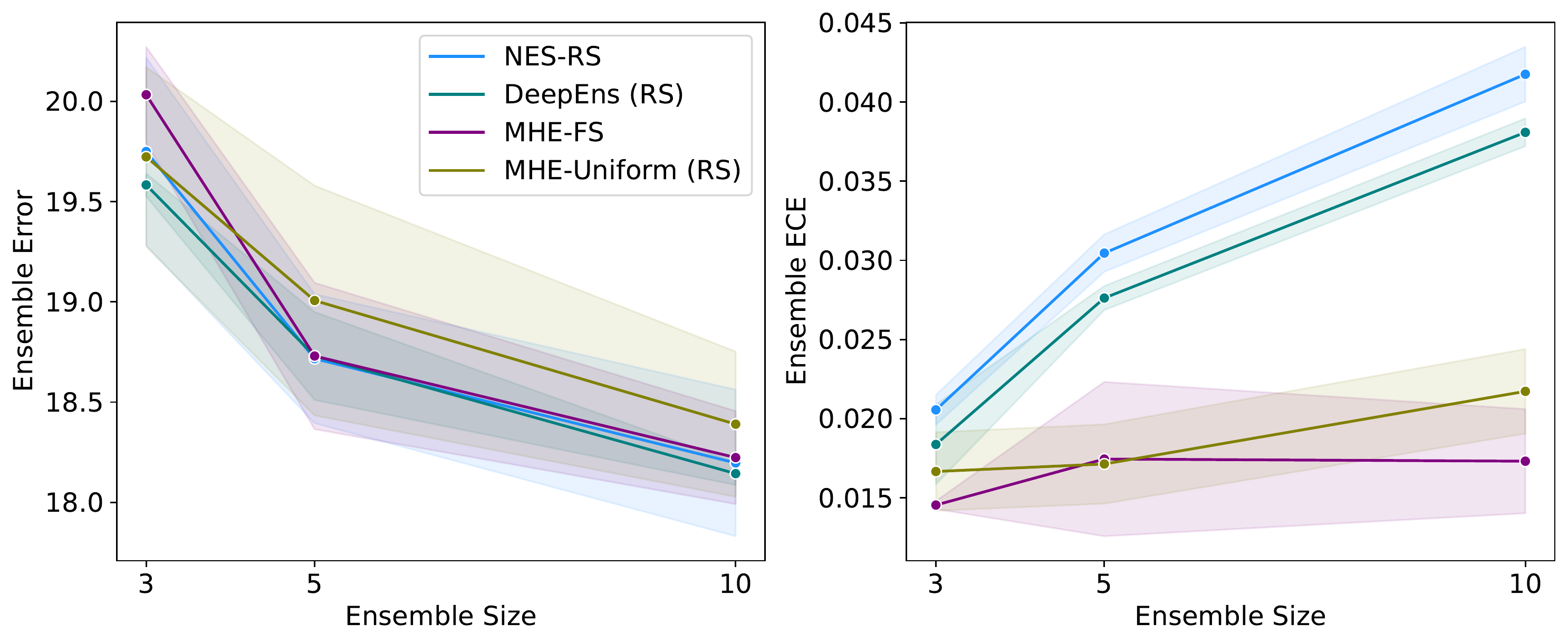}
    \caption{In-distribution performance of independent and multi-headed ensembles on CIFAR-100 dataset. MHE-Uniform (RS) and DeepEns (RS) have same uniform ensemble configuration. NES-RS and MHE-FS have the same ensemble configuration with varying member architectures.}
    \label{fig:appendix:mhe_vs_ind}
\end{figure}

Figure \ref{fig:appendix:mhe_vs_ind} compares the ensemble error and ECE of deep ensembles and multi-headed ensembles on CIFAR-100 for increasing ensemble sizes. 
Both MHE methods are better calibrated compared to deep ensembles, which is clearly evident for larger values of $M$.
For $M=3$, MHE with varying configurations does not perform as well as the other methods. A good uniform configuration can perform better than varying configurations for smaller ensembles, even for deep ensembles.
The effect of varying architectures shows only for medium to large ensembles, when $M \geq 5$.

\clearpage

\end{document}